# Word representation or word embedding in Persian texts

Siamak Sarmady

Erfan Rahmani

(Abstract) Text processing is one of the sub-branches of natural language processing. Recently, the use of machine learning and neural networks methods has been given greater consideration. For this reason, the representation of words has become very important. This article is about word representation or converting words into vectors in Persian text. In this research GloVe, CBOW and skip-gram methods are updated to produce embedded vectors for Persian words. In order to train a neural networks, Bijankhan corpus, Hamshahri corpus and UPEC corpus have been compound and used. Finally, we have 342,362 words that obtained vectors in all three models for this words. These vectors have many usage for Persian natural language processing.

## 1. INTRODUCTION

The Persian language is a Hindi-European language where the ordering of the words in the sentence is subject, object, verb (SOV). According to various structures in Persian language, there are challenges that cannot be found in languages such as English [1]. Morphologically the Persian is one of the hardest languages for the purposes of processing and programing. In addition to common problems with other languages such as ambiguities and distinct combinations, Persian has other challenges. In Persian we have two different type of space character that have different Unicode encodings, The white space that designates word boundaries and pseudo-space that using in compound words [2].

Words on a computer can be stored as strings, but the operation on strings are very inefficient. Computers can manage numbers much better than strings. Here word-to-vector conversion methods have been introduced and the performance of each one is expressed.

When a word is basically stored as a string on a computer, this does not say anything about the concept of the word, and only allows the detection of specific similarities in particular circumstances (For example, when a suffix is added to a word like cats). Of course, it should be noted that many of the words that are conceptually similar do not have the same characters. In addition, doing operations on strings is very inefficient. Better representations also require faster computing that's why now using vector representation for words [3].

In the second section, a variety of word-to-vector conversion methods are presented. In the third section, the models used are introduced. In fourth section, the article methodology is described. In the last section the conclusion is expressed.

## 2. METHODS OF CONVERTING WORD TO VECTOR

The first method of converting a word into the vector was the one-hot vector method. When an application processes text, it can place a word with an index in the vocabulary (E.g. cat, 12424). This is a compact representation for a word. These numbers can be converted to a vector V next to zero in the whole cell, and a bit set in the word index. One-hot vector method misses the similarity between strings that are specified by their characters [3].

The second method is distributed vectors. The words in the same context (neighboring words) may have similar meanings (For example, lift and elevator both seem to be accompanied by locks such as up, down, buildings, floors, and stairs). This idea influences the representation of words using their context. Suppose that each word is represented by a v dimensional vector. Each cell of the vector is related to the one of the words in the vocabulary. Usually, the size of the vocabulary is very large (at least a few hundred thousand words), and working with vectors with these dimensions involves computational obstacles [3]. The third method is the word embedding which is explained in the next section.

## 3. WORD EMBEDDING

The basic concept of words embedding is to store context information in a small-sized vector. Each word is represented by a D-dimensional vector where D is a small value (Usually between 50 and 1000). Instead of counting the number of repetitions of neighbor words, vectors are learned. This work was first performed in 2003 by Bengio et al [4]. The work of Tomas Mikolov et al. [5] and Jeffrey Pennington et al. [6] is the best done, and this two papers are used in this study.

### 3.1. CBOW model

As shown in Figure 1, this model is a language model in which we have a three-layer neural network. Take a few words before and a few words after a word and in output assign a probability score for each of the words in the vocabulary as the probability that the word will appear at the position w (t) (current word). Actually, its purpose is to estimate the current word according to the last few words and the next few words [5].

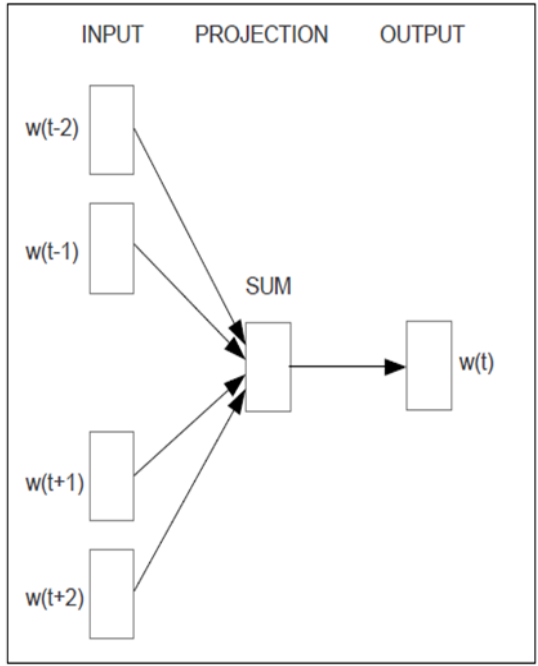

Figure 1: CBOW model [5]

### 3.2. Skip-gram model

As shown in Figure 2, the architecture of this model is similar to CBOW, with the difference that according to the current word, estimates the last few words and the next few words [5].

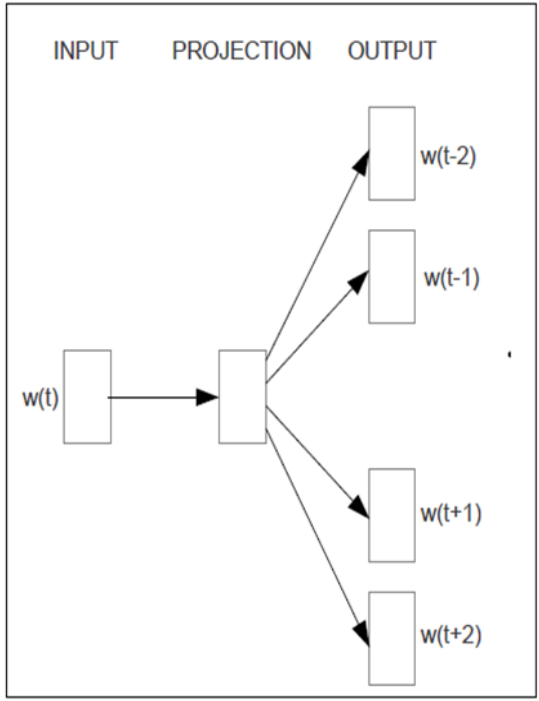

Figure 2: skip-gram model [5]

### 3.3. GLoVe model

This method has a term-term occurrence matrix called X. In which $X_{ij}$ is the number of repetitions of the word i next to the word j, $P_{ij}$ shows the probability of this occurrence. The idea behind this model is to look at the similarity of the two words with respect to the number of repetitions of each of these two words alongside the third word of the text. This is done by the $P_{ik} / P_{jk}$ rate. k is a word of the context. So the purpose of the paper is to obtain the word vectors and the function F, which is by giving the vectors to the F function obtain the rate that specified above. This means that we have:

$$F(w_i, w_j, w_k) = p_{ik} / p_{jk} \qquad (1)$$

Where w is a vector of words. Finally, with respect to this function, a cost function J is defined which aims to minimize it and By minimizing this cost function, optimal vectors are obtained for words. Cost function J is:

$$J = \sum_{i,j=1}^{v} f(x_{ij})(w_i^T w_j + b_i + b_j - \log^{X_{ij}})^2 \qquad (2)$$

Where W is the vector of words and f is a weight function. The weighting function assigns a weight to rare events and events that occur a lot.

$$f(x) = \begin{cases} (x/x_{max})^a & if\ (x < x_{max}) \\ 1(otherwise) \end{cases} \qquad (3)$$

In the end, to test the quality of the vectors, experiments were carried out on comparing words, similarities of words, and NER using GloVe vectors, and the results showed that the GloVe vectors performed better than others [6].

## 4. RESEARCH METHODOLOGY

In this section, the methods that used in this article are presented. First, the corpuses that used in this article and the operations performed on them are described. Then, details of the implementation of the three methods of CBOW, skip-gram and GLoVe, and the preparation of embedded vectors for Persian words are expressed.

### 4.1. Corpuses and preprocessing

In this paper, Bijankhan corpus [7], Hamshahri corpus [8] and UPEC corpus [2] has been used. In the UPEC corpus, the rules of white space and pseudo-space are observed. But there is no pseudo-space in the BijanKhan corpus and Hamshahri corpus. For example, a word such as "کتاب‌ها" in the UPEC corpus has been written with pseudo-space and this is correctness. But in the BijanKhan and Hamshahri, they are written in two forms: "کتابها" and "کتاب ها". In this paper, all three corpuses are combined to learn the algorithms of converting words into vectors. To combine these three corpuses, it was necessary to convert all three corpuses into a simple text format.

3Bijankhan's corpus is available in the form of files in the LBL format that each line has a word and its tag (Figure 3). In order to prepare Bijankhan's corpus in this study, a program has been prepared. This program extracts text from LBL files, Deletes the tags of words, And at the end, it examines pseudo-space. To correct pseudo-space, the spacing character has become pseudo-space in terms like "کتاب ها". But there is no change in the places where the same word was brought in the form of "کتابها". At the end, the corpus is converted from the form of each word in a single line to the normal text and stored in a text file.

The UPEC corpus has the same format as the Bijankhan corpus (Each word and tag in a single line such as Figure 3) but the entire corpus is available as a txt file. To be able to use this corpus to generate embedded vector, we need to convert this corpus to normal text same as Bijankhan corpus. In this research, a program is written for the UPEC corpus that eliminates the labels and converts the whole corpus into normal text.

Figure 3: shape of Bijankhan corpus and UPEC corpus

The Hamshahri corpus is available in 39,411 files in the form of a documentary on the Internet. These documented files are provided in html format.

Figure 4: shape of Hamshahri corpus documents

As seen in the Figure 4, this corpus is written in the form of html files on the Internet. First, a crawler was created and corpus files have been downloaded (39,411 files). In this corpus, there is no pseudo-space. Unlike two other corpus, this corpus has normal format for text and maybe the marks (For example a question mark) adhere to the words and tokenization are not properly.

According to the specifications of the Hamshahri corpus, a program has been prepared for this research that first extracts text from html files then a group of repetitive suffixes and prefixes that are listed in the Table 1 will become pseudo-space. For example, when the word "می رود" with the space written, it becomes pseudo-space ("می‌رود"). But in cases where the word is written as "میرود", there is no change in the word.

Table 1: Repetitive suffixes and prefixes

| example | Suffix or prefix | example | Suffix or prefix |
|---|---|---|---|
| رفته‌ایم | "ایم" | می‌رود | "می " |
| رفته‌اید | "اید" | نمی‌رود | "نمی" |
| رفته‌اند | "اند" | کتاب‌ها | "ها" |
| مجموعه‌ی | "ی" | کتاب‌های | "های" |
| سریع‌تر | "تر" | کتاب‌هایی | "هایی" |
| سریع‌ترین | "ترین" | رفته‌ام | "ام" |
|  |  | رفته‌ای | "ای" |

Next, the redundant characters that may be attached to the words are separated. These symbols are shown in the Table 2.

4Table 2: Repetitive marks

| mark | mark | mark |
|------|------|------|
| ()   | «»   | •    |
| /    | ؛    | '    |
| =    | -    | :    |
| %    | …    | ؟    |
|      | []   | !    |

In the end, all html files and their text are combined and stored in a text file. At this point, the text obtained is ready to use to generate words vector.

### 4.2. Preparation of GLoVe vectors

The implementation code of GLoVe in the Python programming environment is available and called tf-glove [9]. These codes use the TensorFlow library. TensorFlow is an open source software library for machine learning that It is used to train neural networks and is provided by Google [10].

In this research, we have written a program that uses tf-glove codes and use the TensorFlow machine learning library to prepare Persian language vectors. In this program, the data is received as a list of sentences in which the words of each sentence are separated. For this purpose, the Persian text is read from the corpus and the sentences are separated by a tool that is in the NLTK library and they are stored in a list of sentences, and in the next step, the words related to each sentence in the list are tokenized.

The word tokenization functions provided in the Python NLTK Library create problems for Persian words and the pseudo-space character is considered as a word by itself. The split function of the Python can be used to separate words with space. In this case, the function takes words that have pseudo-space correctly as a single word. So after modifying pseudo-space, the same function is used to tokenize words. After completing these steps, the data is ready for presentation to the tf-glove library.

The word "UNK" with "0" id (for anonymous words) and the word "UNK_PAD" with the "1" id (for padding operations in different neural networks) has also been added to the vocabulary.

After learning the word vectors, two files are created as output. The first file is a vocabulary containing words and their indexes (Figure 5A), and the second file is the word vectors, in which each vector is identified by the word id (Figure 5B).

Figure 5: (A) - Part of the vocabulary file (B) - Part of the word vectors

In Table 3, the parameters used in the learning process and implementation of this model are presented.

Table 3: GLoVe Model Parameters

| describtion | value | parameters |
|---|---|---|
| Number of training batchs | 64 | Batch_size |
| Dimensions of word vectors | 128 | Embedding_size |
| The number of considered words in left and right for target word | 5 | Context_size |
| Lowest repeat for a word | 1 | Min_occurences |
| in equation .. x_max Most repeated for each word, | 100 | Cooccourence_cap |
| Learning rate | 0.05 | Learning_rate |
| Alpha value in equation .. | 3/4 | Scaling_factor |
| Number of epoch | 20 | Num_epoch |

### 4.3. Preparation of CBOW vectors

In this study, the CBOW model in the Python environment and using the TensorFlow library, is used [11]. This implementation is aimed at generating appropriate embedded vectors and examining the semantic and syntactic similarity of words in these vectors [5]. But in this research, the goal is to extract word vectors (the weights of this neural network). Due to the difference in the display of Persian words and English words, there are irregularities in Persian display of words. In this implementation, as in the implementation of GLoVe, words are stored as a vocabulary (identified by numeric indexes) in a file. And word vectors are also stored in another file (with the same characteristic indexes). Depending on the word index in the vocabulary, you can find the vector of each word in the vectors file. In this program, the first text of the corpus is read from the input file then the words are tokenized and the number of repetitions of each word is counted and the words are placed in descending order in their vocabulary according to the number of repetitions.

For anonymous words, the words that may later be seen, but not in the vocabulary the word "UNK" is used.

For padding, the word "UNK_PAD" has been added to the vocabulary.

For word tokenization in Persian like the previous section, after correcting pseudo-space in the corpus, the split function in python is used which works correctly with pseudo-space. As stated in the introduction of the CBOW model, this code predicts the target word with respect to a few words before and a few words next (context words). After the end of learning, the weights of the neural network represent the vector of words. After executing this program, word vectors and vocabulary are stored in two separate files. For example, let's consider the following sentences:

"the quick brown fox jumped over the lazy dog"

If the two words are "the" and "brown", or "quick" and "fox" and we want to predict the middle word, the data is stored as follows.

Batch = [ [the , brown] , [quick , fox] , …]

And the target words fall into the same category.

Predict = [ [quick] , [brown] , …]

In Table 4, we observe the important parameters used in the production of vectors embedded by the CBOW model.

Table 4: CBOW model parameters

| describtion | value | parameters |
|---|---|---|
| Number of training batch | 128 | Batch_size |
| Dimensions of word vectors | 128 | Embedding_size |
| The number of considered words in left and right for target word | 1 | Skip_window |
| Number of epoch | 100000 | Num_step |

## 4.4. Preparation of skip-gram vectors

The skip-gram model is very similar to the CBOW model. With the difference that in this model, according to the current word, a few words before target word and a few words next the target word estimated.

The implementation of this model is available as well as CBOW in Python and using TensorFlow [12]. In this code, weighs of neural network after the end of learning show the word vectors. In this study, for the program, as well as the CBOW program, changes have been made to save vectors and vocabulary. The output of this program also includes vocabulary and words vector files. Here the data format that used in this model is shown with an example. Consider the following sentence:

"the quick brown fox jumped over the lazy dog"

If the context window length is equal to 3 in which the target word in the middle and the words in the left and right fields are located, the data set will be as follows:

[quick,[the,brown]]

[brown,[quick,fox]]

…

The goal is to predict the " the " and 'brown" words by knowing the word "quick" as well as predicting the words "quick" and "fox" by knowing the word "brown".

Then the data is converted into inputs and outputs to enter the neural network:

(quick , the)

(quick , brown)

(brown , quick)

(brown , fox)

The important parameters used to generate skip grams embedded vectors are quite similar to the CBOW model and its table.

## 5. CONCLUSION

In this article, the production of embedded vectors for Persian words was discussed. For the production of word vectors, the corpus from the combination of BijanKhan corpus, Hamshahri corpus and UPEC corpus has been used that has a length of 18,000,000 words. The vocabulary contains 342,364 words. And word vectors are derived from three GLoVe, CBOW and skip-gram models that have been updated to work with Persian text. All implementations produce two output files. The first file contains a vocabulary that each of the words in it is characterized by an index. The second file provides the embedded word vectors. Each vector is identified by its corresponding word index (in the vocabulary). These embedded word vectors can be used for many Persian language processing projects such as NER tagging and POS-tagging.